\documentclass[journal]{IEEEtran}

\usepackage{graphicx}
\usepackage{cite}
\usepackage{picinpar}
\usepackage{amsmath}
\usepackage{url}
\usepackage{flushend}
\usepackage[latin1]{inputenc}
\usepackage{colortbl}
\usepackage{soul}
\usepackage{multirow}
\usepackage{pifont}
\usepackage{color}
\usepackage{alltt}
\usepackage[hidelinks]{hyperref}
\usepackage{enumerate}
\usepackage{siunitx}
\usepackage{breakurl}
\usepackage{epstopdf}
\usepackage{makecell}
\usepackage{pbox}

\def\BibTeX{{\rm B\kern-.05em{\sc i\kern-.025em b}\kern-.08em
    T\kern-.1667em\lower.7ex\hbox{E}\kern-.125emX}}
\renewcommand{\vec}[1]{\ensuremath{\boldsymbol{#1}}} 
    \usepackage{preamble}

\begin{document}

\title{Device-Free User Authentication, Activity Classification and Tracking using Passive Wi-Fi Sensing: A Deep Learning Based Approach}
\vspace{-5mm}
 \author{\vspace{-10mm}}
\author{Vinoj Jayasundara*, \IEEEmembership{Member, IEEE,} Hirunima Jayasekara*, \IEEEmembership{Member, IEEE,} Tharaka Samarasinghe, \IEEEmembership{Member, IEEE,} Kasun T. Hemachandra \IEEEmembership{Member, IEEE}
\thanks{* Equal Contribution
	
	V. Jayasundara, H. Jayasekara and K.T. Hemachandra are with the Department of Electronic and Telecommunication Engineering, University of Moratuwa,	Sri Lanka (e-mail:
	\{vinojjayasundara, nhirunima\}@gmail.com, kasunh@uom.lk).

	T. Samarasinghe is with the Department of Electronic and Telecommunication Engineering, University of Moratuwa,	Sri Lanka, and the Department of Electronic and Electrical Engineering, University of Melbourne, Australia (e-mail: tharakas@uom.lk).}

}
\markboth{Preprint. Under Review.}%
{Jayasundara \MakeLowercase{\textit{et al.}}: Device-Free User Authentication, Activity Classification and Tracking using Passive Wi-Fi Sensing: A Deep Learning Based Approach}
\vspace{-1cm}
\maketitle

\begin{abstract}
Privacy issues related to video camera feeds have led to a growing need for suitable alternatives that provide functionalities such as user authentication, activity classification and tracking in a noninvasive manner. Existing infrastructure makes Wi-Fi a possible candidate, yet, utilizing traditional signal processing methods to extract information necessary to fully characterize an event by sensing weak ambient Wi-Fi signals is deemed to be challenging. This paper introduces a novel end-to-end deep learning framework that simultaneously predicts the identity, activity and the location of a user to create user profiles similar to the information provided through a video camera. The system is fully autonomous and requires zero user intervention unlike systems that require user-initiated initialization, or a user held transmitting device to facilitate the prediction. The system can also predict the trajectory of the user by predicting the location of a user over consecutive time steps. The performance of the system is evaluated through experiments.
\end{abstract}


\IEEEpeerreviewmaketitle

\section{Introduction}
\IEEEPARstart{A}{part}from the applications related to surveillance and defense, user identification, behaviour analysis, localization and user activity recognition have become increasingly crucial tasks due to the popularity of facilities such as cashierless stores and senior citizen residences. Current state-of-the-art techniques for passive user authentication\cite{metzler2012appearance}, re-identification\cite{han2019can}, activity classification\cite{simonyan2014two} and tracking\cite{kang2003continuous},\cite{yuan2017automatic} are primarily based on video feed analysis.
However, due to concerns on privacy invasion, camera videos are not deemed to be the best choice in many practical applications. Hence, there is a growing need for non-invasive alternatives.

A possible alternative being considered is ambient Wi-Fi signals, which are widely available and easily accessible. In this paper, we introduce a fully autonomous, non invasive, Wi-Fi based alternative, which can carry out user identification, activity recognition and tracking, simultaneously, similar to a video camera feed. In the following subsection, we present the current state-of-the-art on Wi-Fi based solutions and highlight the unique features of our proposed technique compared to available works.
\vspace{-4.1mm}
\subsection{Related Works}
\subsubsection{\textbf{User Authentication}}
Majority of the wireless aided user authentication systems in the literature require the user to carry or wear a device to facilitate the authentication process \cite{isaac2019template,more2018gait,sun2018accelerometer}. A device free method, where the user need not carry a wireless transmitting device for active user sensing, deems more suitable practically. To this end, WiWho\cite{zeng2016wiwho} and Wi-Fi-ID \cite{zhang2016wifi} utilize conventional signal processing techniques to create a gait profile for each user, which is subsequently used for identification. Research focus, however, has recently shifted towards learning based techniques \cite{pokkunuru2018neuralwave,lin2018wiau}, but being able to handle only registered users is consdiered a major limitation in such systems, \textit{e.g.}, NeuralWave\cite{pokkunuru2018neuralwave}. WiAU\cite{lin2018wiau} focuses on system that is robust to unauthorized users via training their model with both authorized and unauthorized user data. However, training a model for limitless potential unauthorized users is infeasible practically. Our system focuses on providing a robust solution for this limitation.  
\vspace{1mm}
\subsubsection{\textbf{Activity Recognition}}
Wireless aided activity recognition is a well studied area in the literature \cite{pu2013whole,wang2016we,yun2017strata,yousefi2017survey}. It has been already established that deep learning based techniques do outperform the conventional signal processing based techniques \cite{pu2013whole,wang2016we,yun2017strata} with regards to activity recognition (see \cite{yousefi2017survey} and references therein). However, the existing deep learning based systems face difficulties in deployment due to them not considering the recurring periods without any activities in their models. Thus, the systems require the user to invoke the system by conducting a predefined action, or a sequence of actions. This limitation is addressed in our work to introduce a fully autonomous system.

\vspace{1mm}
\subsubsection{\textbf{\noindent User Tracking}}
Wi-Fi based localization systems that utilize deep learning approaches are well studied in the literature \cite{wang2016csi,kim2018scalable,jang2018indoor}, 
with device free Wi-Fi based localization attracting considerable research interest \cite{youssef2007challenges,li2017indotrack,adib20143d}. Most existing deep learning based device free localization systems can predict the position of the user out of a set of pre-determined positions \cite{wang2016device}, but lack the ability to continuously output user co-ordinates, which is mandatory for continuous tracking. This is another gap in the literature that will be bridged in our paper.

\vspace{-3mm}
\subsection{Contributions of the Paper}

We consider a distributed single-input-multiple-output (SIMO)
system that consists of a Wi-Fi transmitter, and a multitude of
fully synchronized multi-antenna Wi-Fi receivers, placed in the sensing area. The samples of the received signals are
fed forward to a data concentrator, where channel state information (CSI) related to all Orthogonal Frequency-Division Multiplexing (OFDM) sub carriers are extracted and pre-processed, before feeding them into the deep neural networks. 
The key features of the proposed system are as follows:
\begin{itemize}
\item  The system is self-sustaining, device free, non-invasive, and does not require any user interaction at the system commencement or otherwise, and can be deployed with existing infrastructure.
\item The system consists of a novel black-box technique that produces a standardized annotated vector for authentication, activity recognition and tracking with pre-processed CSI streams as the input for any event. 
\item  With the aid of the three annotations, the system is able to fully characterize an event, similar to a camera video. 
\end{itemize}




State-of-the-art deep learning techniques can be considered to be the key enabler of the proposed system. With the advanced learning capabilities of such techniques, complex mathematical modelling required for the process of interest can be conveniently learned.  
To the best of our knowledge, this is the first attempt at proposing an end-to-end system that predicts all these three in a multi-task manner. Then, to address limitations in available systems, firstly, for authentication, we propose a novel prediction confidence-based thresholding technique to filter out unauthorized users of the system, without the necessity of any training data from them. Secondly, we introduce a \textit{no activity} (NoAc) class to characterize the periods without any activities, which we utilize to make the system fully autonomous. Finally, we propose a novel deep learning based approach for device-free passive continuous user tracking, which enables the system to completely characterize an event similar to a camera video, but in a non-invasive manner.
The performance of the proposed system is evaluated through experiments, and the system achieves accurate results even with only two single antenna Wi-Fi receivers.

Rest of the paper is organized as follows: in Sections \ref{sec:system_overview}, \ref{sec:process} and \ref{sec:networks}, we present the system overview, methodology on data processing, and the proposed deep neural networks, respectively. Subsequently, we discuss our experimental setup in Section \ref{sec:experiment}, followed by results and discussion in Section \ref{sec:results}. Section \ref{sec:conclusion} concludes the paper.



\vspace{-2mm}
\section{System overview } \label{sec:system_overview}
\begin{figure}[t]
  \centering
  \includegraphics[scale=0.16]{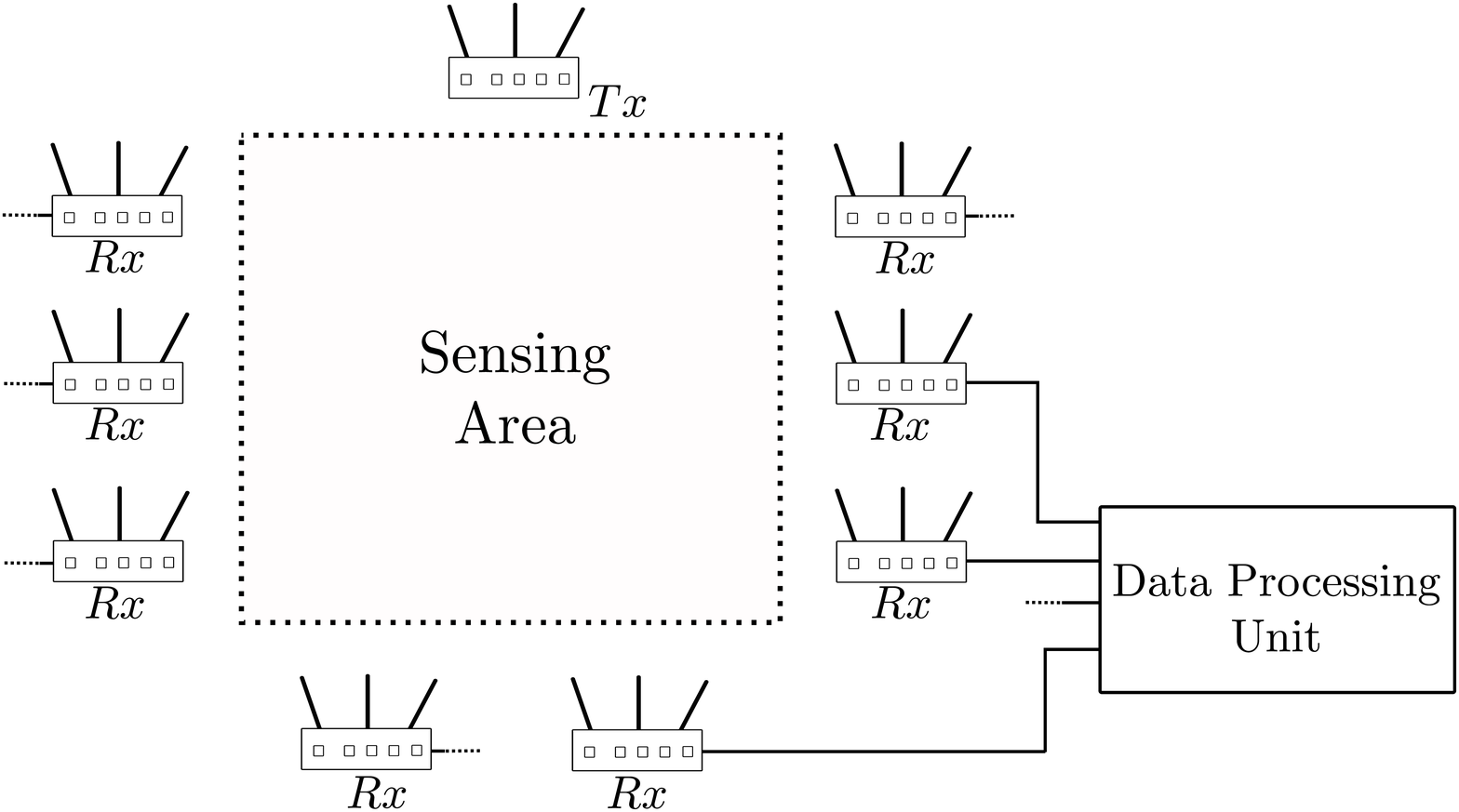}
  \caption{Generic Wi-Fi transmitter and receiver configuration for our system. Here, Rx and Tx refer to receivers and transmitters respectively.}
 \label{fig:system}
  \vspace{-3mm}
\end{figure}



Consider a distributed SIMO system that consists of a single antenna Wi-Fi transmitter, and $M$ Wi-Fi receivers having $N$ antennas each. The transmitter and the receivers are placed in the sensing area, and an example scenario is illustrated in Fig. \ref{fig:system}. The receivers are fully synchronized, with a sampling frequency of $f_s$, and connected to a data concentrator for centralized processing. The received signal at the $n$-th antenna of the $m$-th receiver, where $n \in \{1,\ldots, N \}$ and $m \in \{1,\ldots, M \}$, is given by 
\vspace{-3mm}
\begin{equation} \label{eq:rssi}
\vspace{-2mm}
    y_{m,n}(t) = \sum_{i=1}^S h_{m,n,i,t} \ x_i \cos(2 \pi f_i t+ \theta_{m,n,i,t}) + \eta(t),
\end{equation}
where $S$ is the number of subcarriers in the transmitted OFDM signal. Moreover, for the $i$-th subcarrier, $h_{m,n,i,t}$ and $\theta_{m,n,i,t}$ denote the amplitude and the phase value of the random channel between the transmitter and the $n$-th antenna of the $m$-th receiver, respectively,
and $\eta(t)$ represents the random noise in the received signal. We assume that at a given time $t$, the data concentrator has access to all received signals (samples), 
which can be achieved through a feedforward mechanism.      

\begin{figure}[t]
  \centering
  \includegraphics[scale=0.08]{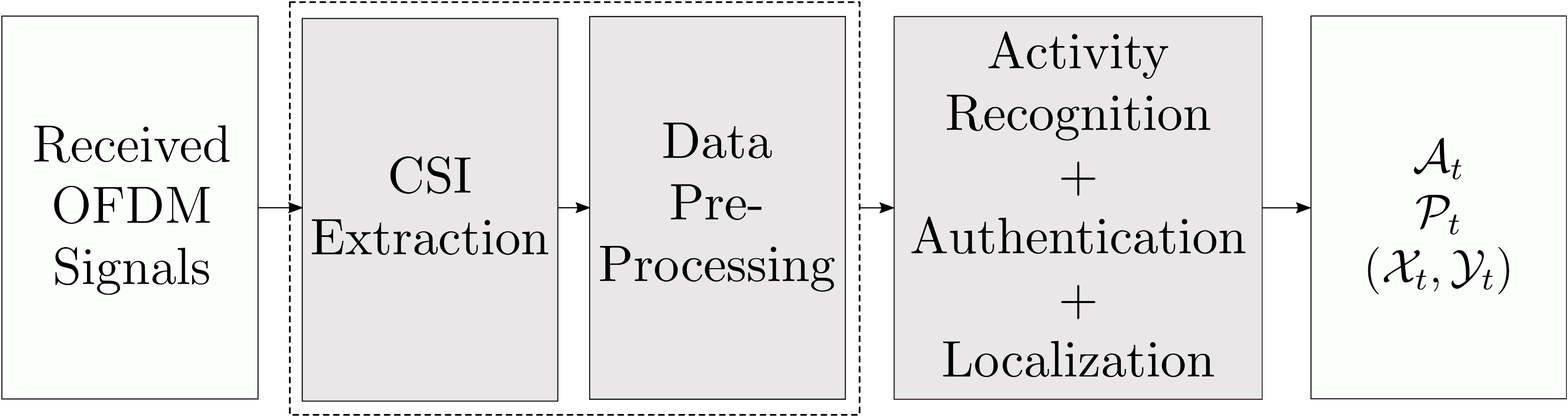}
  \caption{The system architecture. CSI are extracted from the received OFDM signals are pre-processed and fed to the neural networks to obtain $[\mathcal{A}_t,\mathcal{P}_t,(\mathcal{X}_t,\mathcal{Y}_t)]$, where at a given time $t$, $\mathcal{A}_t$ is the predicted activity, $\mathcal{P}_t$ is the predicted person performing the activity and $(\mathcal{X}_t,\mathcal{Y}_t)$ are the Cartesian coordinates of the person's location.}
 \label{fig:overall}
 \vspace{-5mm}
\end{figure}



Fig. \ref{fig:overall} presents an overview of the system.
The first stage, which is implemented at the data concentrator, focuses on extracting CSI from the received signals. CSI is considered to be more stable and robust to complex environmental effects than Received Signal Strength Indicator (RSSI) \cite{li2016robust}, and also since CSI related to each subcarrier of the OFDM can be extracted, the system will have more information for effective learning. Further, how the activities in a sensing area affect the CSI has been recently studied in \cite{zhang2019wifimap+}. This stage is explained in Section \ref{sec:csi}. The second stage is pre-processing the extracted CSI information, which is elaborated in Section \ref{sec:data_preprocess}. The pre-processed data is fed into the deep neural networks, which include the activity recognizer, the authenticator and the tracker. This can be considered to be the third stage in the system architecture, and it is presented in Section \ref{sec:networks}. The deep neural networks output three annotations per data segment of the form $[\mathcal{A}_t,\mathcal{P}_t,(\mathcal{X}_t,\mathcal{Y}_t)]$, where at a given time $t$, $\mathcal{A}_t$ is the predicted activity, $\mathcal{P}_t$ is the predicted person performing the activity and $(\mathcal{X}_t,\mathcal{Y}_t)$ are the Cartesian coordinates of the person's location, relative to a pre-defined coordinate frame. 
With the aid of the three annotations, we can sufficiently characterize the proceedings within a $T$ seconds window, similar to a camera video. 
\section{Data Processing} \label{sec:process}


\subsection{Extracting Channel State Information (CSI)} \label{sec:csi}


Current OFDM implementations (including 802.11a,g,n and ac) use the information available in pilot sub-carriers of the OFDM signal to estimate the channel behaviour and multi-path disturbances caused by the environment. Channel estimation in OFDM systems is well studied in the literature \cite{li2006orthogonal}. 
The data concentrator performs the channel estimation in the proposed system. We note that the channel estimation can alternatively be performed at the receivers, and the estimated CSI can then be forwarded to the data concentrator for further processing. However, most commercial Wi-Fi devices do not provide access to the estimated CSI data, and hence,  we propose a more general architecture for wider adaptability.



Both amplitude and phase values are finer-grained descriptors of the wireless channel \cite{yang2013rssi}. However, the phase values are affected by several sources of error, including the carrier frequency offset (CFO) and the sampling frequency offset (SFO) \cite{yousefi2017survey}. Although these errors can be eliminated using a calibration technique termed data sanitization in \cite{sen2012you}, we avoid phase values in the learning process to reduce the computational and implementational complexity. Thus, at a given time $t$, for each $m \in \{1,\ldots, M \}$ and $n \in \{1,\ldots, N \}$, the data concentrator estimates $\vec{h}_{m,n,t}= [h_{m,n,1,t}\ \ldots\  h_{m,n,S,t}]^\top$, which is a $S$-by-$1$ vector that consists of amplitude values of the channel between the transmitter and the $n$-th antenna of the $m$-th receiver. The concentrator then forwards the estimated amplitude values to the next stage for further processing. 

\vspace{-2mm}
\subsection{CSI Preprocessing} \label{sec:data_preprocess}

We start the CSI preprocessing by a sparsity reduction operation that aids the learning of the network. The coarse frequency offset correction done in channel estimation usually leads to some elements in $\vec{h}_{m,n,t}$ to have negligibly small values (quantitatively, amplitude $<$ 0.05), as evident from Fig. \ref{fig:csi_vs_sub}, irrespective of the trial, activity or the person.  Further, it was noted that this observation held irrespective of the application, distance between the Wi-Fi devices, position of the person, etc. These elements do not provide any useful information and in fact hinders learning. Hence, they are removed from the estimated channel amplitude vectors. We denote the respective sparsity reduced amplitude vector of $\vec{h}_{m,n,t}$ by $\vec{\hat{h}}_{m,n,t}$. As an example, if $\vec{h}_{1,1,t} = [0.1 \ 0.2 \ 0.0 \ 0.4 \ 0.0003]^\top$, we have $\vec{\hat{h}}_{1,1,t}=[0.1 \ 0.2 \ 0.4]^\top$. Moreover, $\vec{\hat{h}}_{m,n,t}$ is a $\bar{S}$-by-$1$ vector, where $\Bar{S} \leq S$. Note that for a given $m$ and $n$, the dimensionality of $\vec{\hat{h}}_{m,n,t}$ is fixed at $\bar{S}$ for all values $t$ since the sparsity is caused by the coarse frequency offset correction. Next, for each $m \in \{1,\ldots, M \}$ and $n \in \{1,\ldots, N \}$, we concatenate the sparsity reduced amplitude vectors along the temporal axis to produce the $\Bar{S}$-by-$T_{jk}$ matrix $\vec{\hat{C}}_{m,n} = [\vec{\hat{h}}_{m,n,1} \ \cdots \  \vec{\hat{h}}_{m,n,T_{jk}}]$. Here, $T_{jk}$ denotes the duration of the $j$-th trial of the $k$-th activity, signifying the variability in the duration of activities, as well as the variability in the duration of different trials of the same activity.

Next, we focus on noise removal. In order to reduce burst noise, we carry out Butterworth filtering on each time series represented by the rows of $\vec{\hat{C}}_{m,n}$. 
Even though principal component analysis (PCA) based noise removal is proven to be more effective at burst noise removal in CSI signals than low-pass filtering \cite{wang2015understanding}, we again resort to the low complex method to minimize the implementation complexity. Butterworth low pass filtering provides sufficiently adequate noise removal, as shown later in our experimental results. We denote the noise filtered data matrix of $\vec{\hat{C}}_{m,n}$ by $\vec{\tilde{C}}_{m,n}$. Guidelines on selecting the cutoff frequency of the low pass filter are also provided with reference to the experiment in the section \ref{sec:experiment}.  




We concatenate all filtered CSI for a particular activity such that the resultant $MN\Bar{S}$-by-$T_{jk}$ matrix is given by $\vec{\Tilde{C}}= [\vec{\tilde{C}}_{1} \cdots \vec{\Tilde{C}}_{M}]^\top$, where $\vec{\tilde{C}}_{m} = [\vec{\tilde{C}}_{m,1} \cdots \vec{\Tilde{C}}_{m,N}]^\top$ for all $m \in \{1,\ldots,M\}$. 
We segment $\vec{\Tilde{C}}$ into time steps of $f_sT$ samples (corresponding to $T$ seconds) with 90\% overlap, to produce the labelled dataset $D$ of the dimensions $[R,f_sT,MN\Bar{S}]$, where $R$ is the number of training/testing samples. 
It is necessary to maintain that $T<T_{jk}, \ \forall j,k$, such that $T$ will be less than the lowest duration of any trial of any activity. 
The whole data processing stage is summarized in Algorithm \ref{algo:1}.
\renewcommand{\algorithmicrequire}{\textbf{Input:}}
\renewcommand{\algorithmicensure}{\textbf{Output:}}

\begin{algorithm}[!h] \caption{CSI data pre-processing Algorithm} \label{algo:1}
\algorithmicrequire \ $y_{m,n}(t) \ \forall m \in [1, ... ,M]$, $\forall n \in [1, ... ,N]$ and $\forall t$ \newline
\algorithmicensure \ Processed dataset $D$ to be fed in to the deep networks

\vspace{-1mm}
\begin{algorithmic}[1]
\STATE $ \forall m,n, \vec{h}_{m,n,t} \gets Estimate \ Channel(y_{m,n}(t))$
\STATE $\forall m,n, \vec{\hat{h}}_{m,n,t} \gets  Sparsity \ Reduce(\vec{h}_{m,n,t})$
\STATE $\forall m,n, \vec{\hat{C}}_{m,n} \gets [\vec{\hat{h}}_{m,n,1} \ \cdots \  \vec{\hat{h}}_{m,n,T_{jk}}]$
\STATE $\forall m,n, \vec{\Tilde{C}}_{m,n} \gets Butterworth(\vec{\hat{C}}_{m,n})$ 
\STATE $\forall m, \vec{\Tilde{C}_m} \gets [\vec{\Tilde{C}}_{m,1} \cdots \vec{\Tilde{C}}_{m,N}]^\top$
\STATE $ \vec{\Tilde{C}} \gets [\vec{\Tilde{C}}_{1} \cdots \vec{\Tilde{C}}_{M}]^\top$
\STATE $D \gets Segment(\vec{\Tilde{C}})$
\end{algorithmic}
\end{algorithm}
\vspace{-5mm}
\section{Deep Networks for Activity Classification, Authentication and Tracking} \label{sec:networks}

\begin{figure}[t]
  \centering
  \includegraphics[scale=0.1]{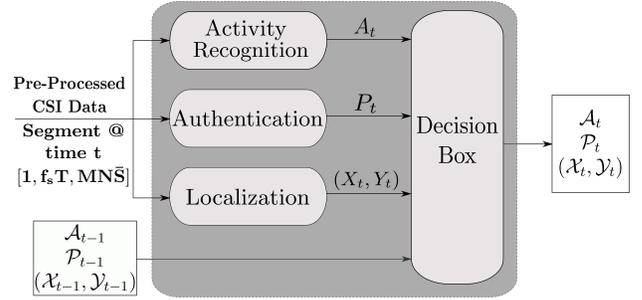}
  \caption{Generating annotations using deep neural networks. During the inference phase, the decision box aids to distinguish between static and dynamic activities, to predict $[\mathcal{A}_t,\mathcal{P}_t,(\mathcal{X}_t,\mathcal{Y}_t)]$ correctly, where at a given time $t$, $\mathcal{A}_t$ is the predicted activity, $\mathcal{P}_t$ is the predicted person performing the activity and $(\mathcal{X}_t,\mathcal{Y}_t)$ are the Cartesian coordinates of the person's location.}
 \label{fig:overall_networks}
 \vspace{-5mm}
\end{figure}

We consider a threefold classification of activities as \textit{dynamic}, \textit{static} and \textit{no activity} (NoAc). Firstly, we classify an activity as dynamic if the location coordinates of the performer varies significantly during the action. This includes activities such as running and walking. Secondly, we classify an activity as static if the location coordinates of the performer do not vary significantly during the action. For example, activities such as sitting, jumping and falling can be categorized as static activities if minor location coordinate changes are disregarded. Finally, as a novel and very important contribution, we classify scenarios where there is no one in the sensing area, or the performer in the sensing area is not engaging in any activity, as NoAc. If NoAc is not captured in the classification process, the system will require the user to initiate the system before performing the activity, \textit{e.g.}, through a push button input. Classification of NoAc makes this a system that requires zero user interaction. 

\renewcommand{\algorithmicrequire}{\textbf{Input:}}
\renewcommand{\algorithmicensure}{\textbf{Output:}}



\vspace{-2mm}
\subsection{Activity classification} \label{sec:activity}
\begin{figure}[t]
    \vspace{-2mm}
  \centering
  \includegraphics[scale=0.8]{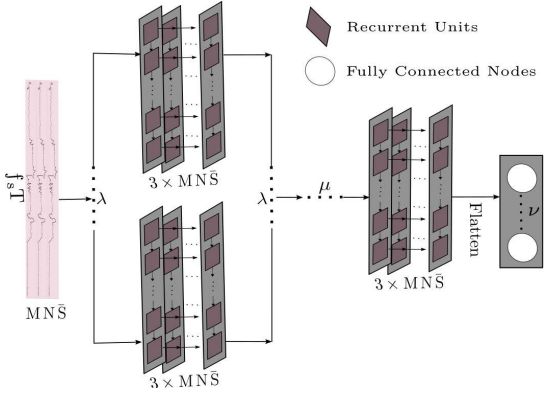}
  \caption{General Network Architecture. Let $\lambda$ be the number of parallel hidden layers, $\mu$ be number of sequential hidden layers after the block of parallel layers and $\nu$ be the dimension of the output vector for each task. For the activity recognizer, $\lambda = 1 , \mu = 1, \nu = K$, for the authenticator $\lambda = 1 , \mu = 2, \nu = P$, and for the tracker $\lambda = 2 , \mu = 1, \nu = 2$. Here, $M$ is the number of receivers, $N$ is the number of antennas per receiver and $\Bar{S}$ is the number of subcarriers after sparsity reduction.}
 \label{fig:act_net}
 \vspace{-1mm}
\end{figure}

We propose the deep neural network illustrated in Fig. \ref{fig:act_net} for activity classification. The network consists of two recurrent layers with a dropout of 0.3 \cite{srivastava2014dropout} in between to reduce overfitting, followed by one fully connected layer. We use three different types of recurrent layers in our study, Long Short-Term Memory (LSTM), Gated Recurrent Unit (GRU) and Bidirectional-Gated Recurrent Unit (B-GRU) in our experiments in an attempt to find the best suited recurrent layer type for each task. Following the recommendation of \cite{srivastava2014dropout}, it is widely accepted that, in order to avoid overfitting, dropout should be used prior to the layers with the highest percentage of trainable parameters resulting them more prone to co-adapt them self to training data. For our networks, the highest percentage of the trainable parameters (more than $75\%$) are concatenated in the recurrent layers. Hence, we add dropout layers between recurrent layers. Literature suggests that in general, the dropout rate is set between 0.2 \cite{huang2017densely} and 0.5\cite{srivastava2014dropout} to achieve optimum results. Hence, in compliance, we empirically set the drop rate to 0.3 for our application. The two recurrent layers have tanh activation in order to maintain the layer outputs within $(1,-1)$, similar to the normalized CSI streams \cite{graves2013speech}, whereas the final fully connected layer has softmax activation following the existing state-of-the-art classification approaches \cite{nwankpa2018activation}. Each recurrent layer consists of $3 \times MN\Bar{S}$ hidden units, and the fully connected layer consists of $K$ units, where $K$ is the number of activity classes, $M$ is the number of receivers, $N$ is the number of antennas per receiver and $\Bar{S}$ is the number of subcarriers after sparsity reduction. Following the existing state-of-the-art classification approaches with more than two distinct classes, we use {categorical cross-entropy} as the loss function, where the total loss for activity recognition is given by
\vspace{-1mm}
\begin{equation} \label{eq:cate_cross_act}
    L_{\text{activity}} = - \frac{1}{R} \sum_{r=1}^{R} \sum_{k=1}^{K} a_{r} \ \log (b_{r,k}),
\end{equation}
where $a_r$ is the ground truth of the $r$-th sample, and $b_{r,k}$ is the predicted value for the $k$-th activity of the $r$-th training/testing sample. The network uses adam with a learning rate of $0.001$ as the optimizer, following \cite{kingma2014adam}.


\vspace{-2mm}
\subsection{User Authentication} \label{sec:authentication}
The network we propose for authentication consists of three recurrent layers with a dropout of 0.3 in between the first and the second layers to reduce overfitting, followed by one fully connected layer.
The fully connected layer consists of $P$ units, where $P$ is the number of participants. The rest of the network parameters are similar to the activity recognizer in \ref{sec:activity}. Using the same loss function as earlier, the total loss for activity recognition is given by
\vspace{-1mm}
\begin{equation} \label{eq:cate_cross_auth}
    L_{\text{auth}} = - \frac{1}{R} \sum_{r=1}^{R} \sum_{p=1}^{P} c_{r} \log (d_{r,p}),
\end{equation}
\vspace{-1mm}
where 
$c_r$ is the ground truth of the $r$-th sample, and $d_{r,p}$ is the predicted value for the $p$-th person of the $r$-th sample. 
\vspace{-2mm}
\subsection{Tracking} \label{sec:localization}
The tracking network consists of two recurrent layers. First layer consists of two parallel recurrent layers with each having $3 \times MN\Bar{S}$ hidden units, which will extract the low level features from the input sequence. Second recurrent layer consists of $3 \times MN\Bar{S}$ hidden units, which identifies the remaining patterns present in the data sequence. Dropout of 0.2 is used in between two layers. Final regression layer is trained to regress on x and y distances using features which were extracted in the second recurrent layer. We use mean squared error (MSE) as the loss function such that the total loss for tracking is given by
\begin{equation} \label{eq:mse}
    L_{\text{tracking}} = \frac{1}{R} \sum_{r=1}^{R} (x_{p,r} - x_{t,r})^2 + \frac{1}{R} \sum_{r=1}^{R} (y_{p,r} - y_{t,r})^2,
\end{equation}

where $(x_{p,r},y_{p,r})$ are the predicted Cartesian coordinates and $(x_{t,r},y_{t,r})$ are the ground truth Cartesian coordinates of the $r$-th training/testing sample.

Proper training is critical for the performance of the system. It is not hard to see that the tracking network should only be trained on dynamic activities because effective learning necessitates significant changes in location. Similarly, the authentication network should only be trained on dynamic activities, as it will not be effective to authenticate the performer after learning through static activities. For example, consider authenticating a performer from the manner in which he jumps or falls. On the other hand, the activity recognition network is trained on all these three categories of activities. In this case, even not doing an activity should be classified, and hence, NoAc is of importance as well. Note that in a practical activity recognition system, NoAc will be the most common and frequent activity. Since the activity labels are available during the training phase, we can conveniently train the authentication and tracking networks with only the dynamic activities and the activity recognition network with all activity classes, including the balanced NoAc class.

Now, let us focus on the inference phase, that is, the operation of a functioning deployed system. In the inference phase, we do not have knowledge on the activity classification as a priori. Therefore, the incoming CSI data is fed into all three networks for prediction, irrespective of the activity. Obviously, the authenticator and the tracker will fail to predict accurately for static and NoAc categories. However, since we have an activity predictor, we can observe the output of the activity predictor, and discard the predictions of the authenticator and the tracker for static and NoAc categories, as illustrated by the decision box in Fig. \ref{fig:overall_networks}. 
We update the activity $\mathcal{A}_t$, authentication $\mathcal{P}_t$ and location $(\mathcal{X}_t,\mathcal{Y}_t)$ annotations if $\mathcal{A}_t$ is a dynamic activity. If not, we discard the predicted authentication and location annotations, and assign them the previously predicted annotations, \textit{i.e.}, $\mathcal{P}_t=\mathcal{P}_{t-1}$ and $(\mathcal{X}_t,\mathcal{Y}_t)=(\mathcal{X}_{t-1},\mathcal{Y}_{t-1})$, respectively. Next, we present the deep neural networks proposed for each task.

\vspace{-2mm}
\subsection{Combined multi-task network for dynamic activities} \label{sec:combined}
\begin{figure}[t]
  \centering
  \includegraphics[scale=0.065]{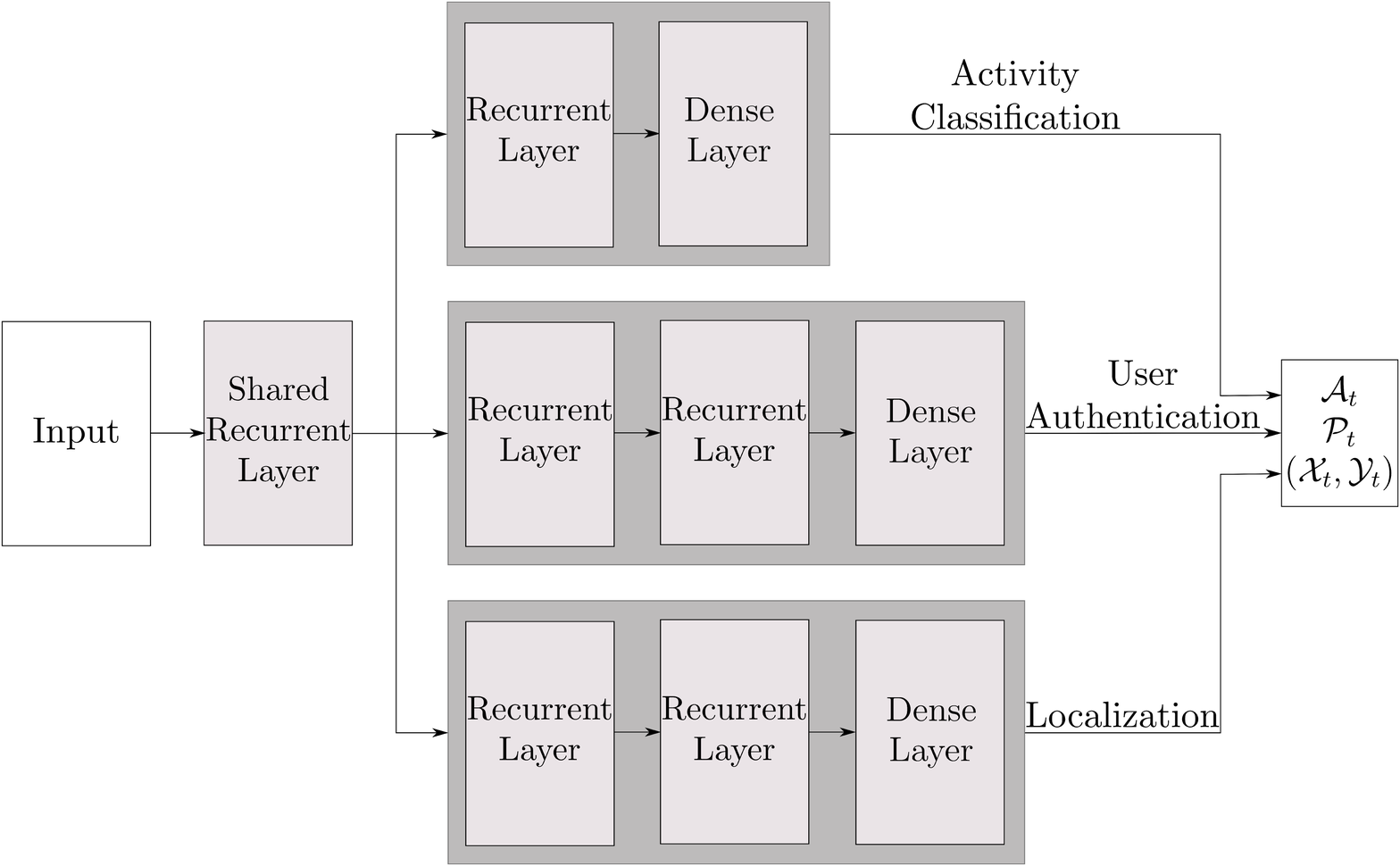}
  \caption{Combined single multi-task network. For the dynamic activities, the pre-processed CSI data can be fed to the shared layer, in order to concurrently generate $[\mathcal{A}_t,\mathcal{P}_t,(\mathcal{X}_t,\mathcal{Y}_t)]$, where at a given time $t$, $\mathcal{A}_t$ is the predicted activity, $\mathcal{P}_t$ is the predicted person performing the activity and $(\mathcal{X}_t,\mathcal{Y}_t)$ are the Cartesian coordinates of the person's location}
 \label{fig:combined}
\end{figure}

It is evident that the authenticator and the tracker should only be trained on dynamic activites, whereas the activity recognizer can be trained on both static and dynamic activities. Due to such varying input datasets necessary for the activity recognizer, authenticator and the tracker, it is challenging to design a single network for all three tasks. However, for applications where only dynamic activities are considered, we propose a single multi-task network, as illustrated by Fig. \ref{fig:combined}.
The network consists of an initial recurrent layer, which is shared by all the three tasks. Lower level features learnt for all three tasks are similar, hence, the initial layer can be jointly learnt by the three tasks. Subsequently, the network splits into three task heads, each corresponding to activity recognition, classification and trackng tasks. Each recurrent layer has tanh activation, and every fully connected layer in the classification heads has softmax activation, whereas every fully connected layer in the regression head has linear activation.

We define the objective function for the proposed multi-task network as a weighted sum of the three distinct loss functions defined in \eqref{eq:cate_cross_act}, \eqref{eq:cate_cross_auth} and \eqref{eq:mse}, which is given by
\begin{equation} \label{eq:combined}
    L^* = \alpha \ L_{\text{activity}} + \beta \ L_{\text{auth}}+ \gamma \ L_{\text{tracking}},
\end{equation}
where $0<\alpha,\beta,\gamma<1$ and $\alpha+\beta+\gamma = 1$. 


\section{Experimental Validation} \label{sec:experiment}
\subsection{Experimental Setup}\label{sec:experiment_setup}
\begin{figure}[t]
  \centering
  \includegraphics[scale=0.2]{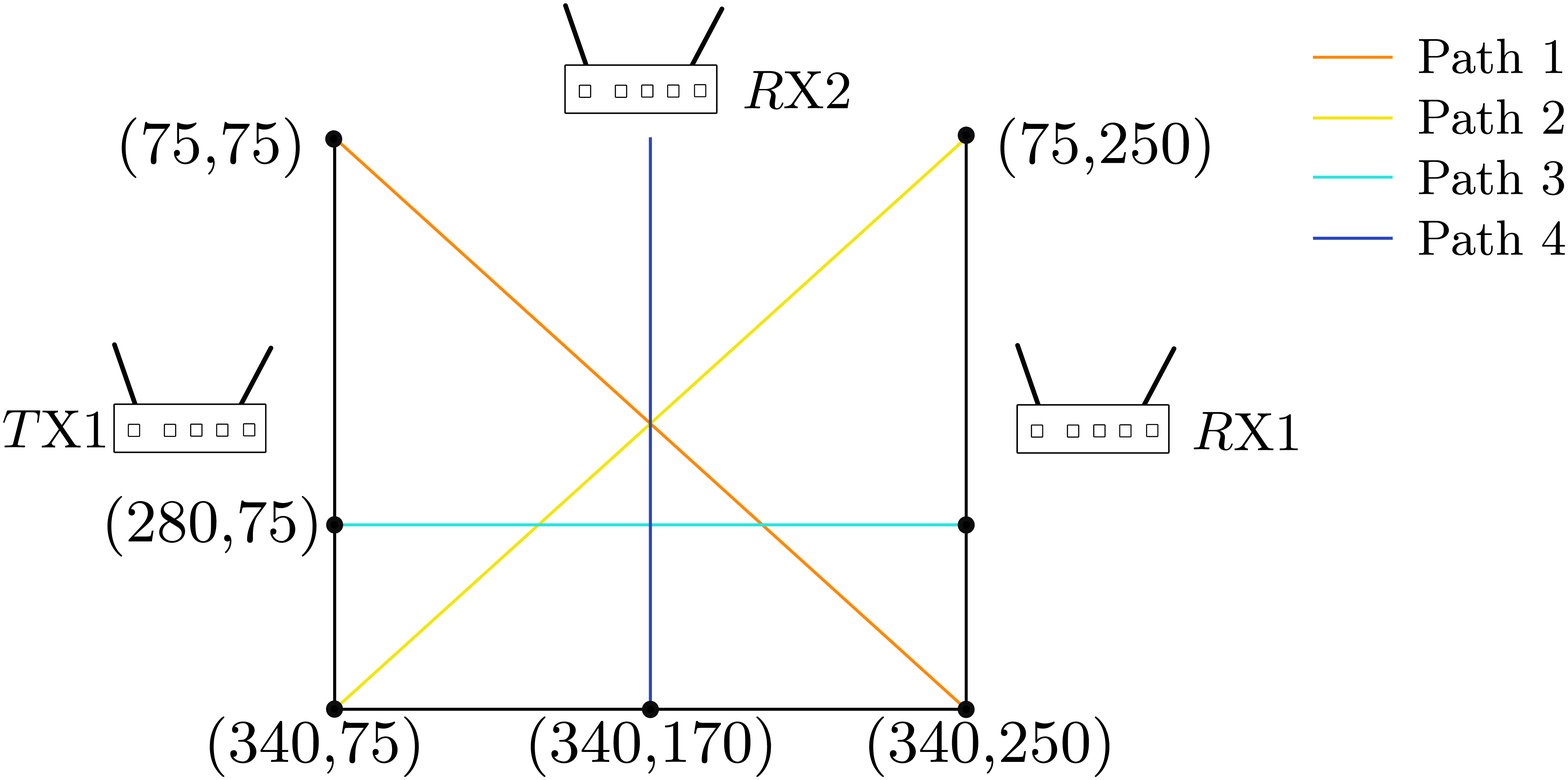}
  \caption{Experiment setup. Different colored lines show the four different walking paths we used to experiment tracking algorithm. Here, TX1, RX1, RX2 refer to the transmitter antenna, receiver antenna directly placed opposite to TX1 and the receiver antenna placed perpendicular to the line that joins TX1 and RX1. The Coordinates are in cm.}
 \label{fig:setup}
 \vspace{-4mm}
\end{figure}

We deploy our system using three 
Universal Software Radio Peripheral (USRP) N210 
software defined radios (SDRs), each configured to have one omni directional antenna, such that we have one transmitter and two receivers. The acquisition sampling rate, $f_s$, for each antenna is 100Hz. 
Optimum placement of the transmitter and the receivers is not studied in this paper, and we resort to the simplest form of placing the two receivers. To this end, one receiver (RX1) is placed directly opposite to the transmitter, and the other (RX2) is placed perpendicular to the line that joins the transmitter and the first receiver (RX1), as illustrated in Fig. \ref{fig:setup}. The perpendicular placement allows us to achieve perspective invariance. We show later that the results obtained from this simple setup, which may or may not be optimal, can be used to get insights on the optimal placement of receivers. 
The SDRs are programmed to transmit and receive Wi-Fi packets with 64 subcarriers ($S = 64$), and in the 2.45GHz frequency band. The SDRs that act as receivers are utilized for the channel estimation. The experiment is done inside an indoor environment (lecture room), where there is rich scattering, and high interference from other Wi-Fi networks and radio frequency (RF) sources. 

\vspace{-2mm}
\subsection{Data Collection}

\begin{figure}[t]
  \centering
  \includegraphics[scale=0.5]{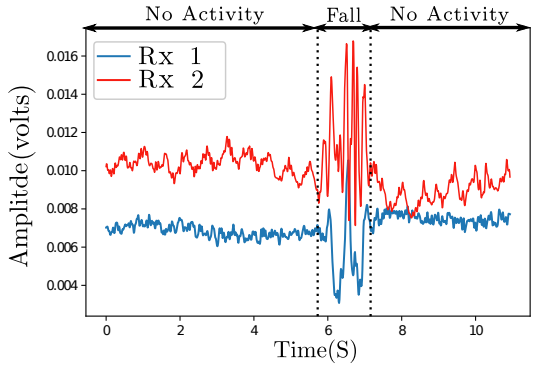}
  \caption{Activity and no-activity labelling of the CSI streams, in order to obtain the training data for each activity class. Here, RX1 and RX2 refer to the two receiver antennas.}
 \label{fig:ac_noac}
\end{figure}

We collect data using $P=13$ voluntary participants, where each participant is requested to perform a set of pre-designated activities. For the experiment, we use three static activities \textit{sit}, \textit{fall} and \textit{jump}, two dynamic activities \textit{walk} and \textit{run}, and the \textit{NoAc} category.
The data collection duration for one activity trial is 10 seconds, during which the participant is instructed to remain stationary for a brief period, prior to commencing the activity, and after completing the activity until instructed to stop. The \textit{walk} activity consists of four different paths (resulting in $K=9$: \textit{sit, jump, fall, run, NoAc, walk1, walk2, walk3, walk4}), which are marked on the floor, as illustrated by Fig \ref{fig:setup}.
We obtain camera video recordings of each trial in order to annotate the collected CSI streams. To this end, the stationary periods are annotated as NoAc and the activity period is annotated with the corresponding activity, as illustrated by Fig. \ref{fig:ac_noac}.


\begin{figure}[t]
  \centering
  \includegraphics[scale=0.36]{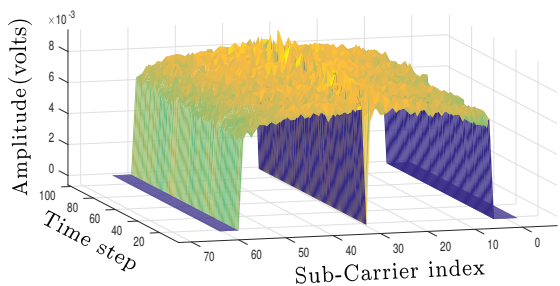}
  \caption{A sample $\vec{\hat{h}}_{1,1,t}$, which is the estimated channel state vector, as a function of the subcarrier index. It can be clearly observed that amplitudes of certain subcarriers are approximately zero.}
 \label{fig:csi_vs_sub}
 \vspace{-4mm}
\end{figure}

Fig. \ref{fig:csi_vs_sub} illustrates an example for an estimated channel state vector, obtained from a commercial Wi-Fi device with $64$ sub-carriers. Note that sub-carriers with indices $[1,2,3,4,5,6,33,60,61,62,63,64$] have approximately zero amplitude in this case, and hence, highlighting the importance of  the sparsity reduction proposed in Section \ref{sec:data_preprocess}. 
Applying the data preprocessing proposed in Section \ref{sec:data_preprocess} on the extracted CSI results in frames of dimensions $[1,80,104]$, for $T=0.8$ s. The duration of the \textit{jump} activity generally registers the lowest duration, between $0.9-1.1$ s. Hence, the value of $T$ is chosen to be $0.8$ s, which is less than $0.9$ s. We assume that human movements typically do not exceed $10$ m/s, which is consistent with the $7.7$ m/s limit set in \cite{wang2015understanding}, where running is not considered. Thus, typical human movements introduce frequencies less than $\frac{10}{0.1223/2} = 163.53 $ Hz \cite{wang2015understanding} at a wavelength of $12.23$ cm since our system operates at $2.45$ GHz. Hence, the cut-off frequency of the $10^{th}$ order Butterworth low pass filter is set to $200$ Hz.

For the classification tasks, which are activity recognition and authentication, we train two different models, that consists of LSTM layers, and GRU layers, respectively. For the regression task, which is tracking, we train three different models, that consists of the state-of-the-art recurrent layers: LSTM layers, GRU layers and B-GRU layers \cite{graves2013speech}, respectively.

A major challenge in generating a dataset for activity recognition is the class imbalance due to the variable durations of the activities. For example, the \textit{sit} activity has an average duration of $1$ s, whereas the \textit{run} activity has an average duration of $4$ s, resulting in more data samples. On average, NoAc accounts for 60\%-90\% of the duration of each trial, largely contributing to the data imbalance. In an attempt to reduce this inherent class imbalance, we vary the number of trials (estimated based on the average duration per activity and the comfort of the participant in performing repeated trials) for each activity inversely proportional to the duration of the activity. Hence, for \textit{sit} and \textit{jump} activities, we conduct 10 trials for each participant, whereas we conduct 8 trials for the \textit{fall} activity and 6 trials for the \textit{run} and \textit{walk} activities, for each participant, respectively. The reduced number of training samples due to the low activity duration can be compensated by obtaining samples from a higher number of trials. Further, we randomly sample the \textit{no activity} duration in order to obtain training samples matching the average number of training samples obtained for other activity classes, thus resolving the class imbalance.

\vspace{-2mm}
\subsection{Robustness analysis of the proposed networks} \label{sec:robust}
We study the robustness of the proposed networks through several experiments. We use the collected data from an in-the-wild participant, whose data is not used for training, to investigate the robustness of the activity recognizer and the authenticator. We hypothesize that if the two networks are sufficiently robust, the activity recognizer should successfully classify the activities with high confidences, whereas the authenticator should fail to provide a prediction with high confidence. 
\vspace{-4mm}
\section{Results and Discussion} \label{sec:results}
\def\colorModel{hsb} 

In this section, we present and discuss the results of the experiments. We use ensembling techniques \cite{dietterich2000ensemble} to improve the performance of our networks by reducing the model variance. We take a simple weighted voting of the predictions of the set of classifiers to produce the ensemble prediction. Each model is trained on a GTX 1080 graphics processing unit (GPU) for 60 epochs. The results obtained for the proposed networks for the three tasks, with different types of recurrent layers are summarized in Table \ref{table:networks}. Each result depicts the average of the maximum accuracy obtained in 3 independent trials. The average inference time per training sample in milliseconds for each model is summarized in Table \ref{table:time}.
\setlength\tabcolsep{1.5pt}
\begin{table}[t]
\caption{Results of the proposed networks for different types of recurrent layers.}
\centering
\begin{tabular}
{|p{1.65cm}|p{0.94cm}|p{0.94cm}|p{0.94cm}|p{1.2cm}|p{1.1cm}|p{0.86cm}|}

\hline
Task &LSTM&GRU& B-GRU &Ensemble Model&Precision&Recall\\
\hline
    Activity recognition&96.36\%& 98.11\%&-&\textbf{98.74\%}&0.9883&0.9826\\
\hline
Authentication & 94.41\% & 95.53\% & - & \textbf{97.41\%}&0.9754&0.9745\\
\hline
Tracking & $\pm 21$ cm & $\pm 19$ cm & $\pm 10$ cm & \textbf{$\pm$10 cm}&-&-\\
\hline
\end{tabular}
\label{table:networks}
\end{table}
\setlength\tabcolsep{3pt}
\begin{table}[t]
\vspace{-4mm}
\caption{Inference time per 800 ms long sample in milliseconds.}
\centering
\begin{tabular}{ |{c}|p{1cm}|p{1cm}|p{1cm}|p{1.3cm}|}
\hline
Task & LSTM & GRU & B-GRU & Ensemble Model\\
\hline
Activity recognition & 4.28 ms & 3.75 ms & - & \textbf{5.82 ms} \\
\hline
Authentication & 6.32 ms &  5.63 ms & - & \textbf{8.64 ms}\\
\hline
Tracking & 3.23 ms & 3.10 ms & 5.60 ms & \textbf{8.72 ms}\\
\hline
\end{tabular}
\label{table:time}
\vspace{-4mm}
\end{table}
\vspace{-3mm}
\subsection{Performance of the Activity Recognition Network}

\newcommand\ColCell[1]{
  \pgfparse{#1<50?1:0}  
    \ifnum\pgfmathresult=0\relax\color{white}\fi
  \pgfmathsetmacro\compA{0}      
  \pgfmathsetmacro\compB{#1/100} 
  \pgfmathsetmacro\compC{1}      
  \edef\x{\noexpand\centering\noexpand\cellcolor[\colorModel]{\compA,\compB,\compC}}\x #1
  } 

\newcolumntype{E}{>{\collectcell\ColCell}c<{\endcollectcell}}

It is evident from Table \ref{table:networks} that the proposed model with GRU layers marginally outperforms the model with LSTM layers (by $1.45\%$). Nevertheless, ensembling over the said two models provides the best performance of 98.74\%, outperforming the GRU model by a margin of $0.63\%$. According to Table \ref{table:time}, the time taken by the ensemble model to perform inference on a sample 800 ms long is 5.82 ms, which is higher than that for the LSTM and GRU models, yet, acceptable for a real time system since inference is completed in $0.72\%$ of the total sample duration. Table \ref{table:error rate} tabulates error rates for each activity used for activity recognition. The highest error rate of $2.49\%$ is recorded by the $3$-rd variation of the \textit{walk} activity, yet, it is within acceptable margins. The \textit{NoAc} class is evidently distinct from every other activity classes and the \textit{jump} activity contains upwards motion not present in any other activity class, resulting in 0\% error rate. Furthermore, \textit{sit} and \textit{fall} have similar downwards motions, and \textit{run} and \textit{walk} have similar forwards motions,causing them to have higher error rates.


\vspace{-2mm}
\subsection{Performance of the Authentication Network}
Similar to the activity recognizer, it can be observed from Table \ref{table:networks} that the proposed model with GRU layers marginally outperforms the model with LSTM layers (by $1.12\%$) and ensembling over the said two models provides the best performance, outperforming the GRU model by a margin of $1.88\%$. According to Table \ref{table:time}, the time taken by the ensemble model to perform inference on a sample 800 ms long is 8.64 ms, which is higher than that for the LSTM and GRU models, yet, acceptable for a real time system since inference is completed in $1.08\%$ of the total sample duration.
\setlength\tabcolsep{4.5pt}
\begin{table}[t]

\caption{Comparison on our Tracking method, with the state-of-the-art device free Authentication results. $P$ transmitters with $Q$ antennas each and $M$ receivers with $N$ antennas each}
\centering
\begin{tabular}{ |p{1.15cm}|p{2cm}|p{1cm}|p{1cm}|p{1cm}|}
\hline
Method & Sensing area(m) & $M\times N$ & $P\times Q$ & Error\\
\hline
Youssef \textit{et al}\cite{youssef2007challenges} & 
$2.74$m long strip &
$2 \times2$ &
$2\times2$ &
$\pm15.7$\textcolor{blue}{cm}\\
\hline

IndoTrack&
$7\text{m}\times7.4$m &
$2\times3$&
$2\times3$&
$\pm62$cm\\
\cite{li2017indotrack}&
$68.5 \text{m}^2$&
$4\times3$&
$2\times3$&
$\pm62$cm\\
\hline
This paper&
$3.4\text{m}\times2.5\text{m}$&
$2\times1$&
$1\times1$&
$\pm10\text{cm}$
\\
\hline
\end{tabular}
\label{table:local_cmp}
\end{table}
\vspace{-4mm}
\subsection{Performance of the Tracking Network}
By referring to Table \ref{table:networks}, the proposed model with GRU layers slightly outperforms the model with LSTM layers by an error margin of $\pm$ 2 cm, which is an improvement of approximately $9.52\%$, whereas the model with B-GRU layers significantly outperforms the model with GRU layers by an error margin of $\pm$ 9 cm, which is a significant improvement of approximately $47.36\%$. It achieves a mean squared error of $\pm$ 10 cm on both $x$ and $y$ coordinates, which is highly acceptable for an indoor tracking system. Table \ref{table:local_cmp} compares our system with the existing state-of-the-art, in terms of the size of the sensing area, number of transmitters/receivers used, mean squared error achieved, etc.

\begin{figure}[t]
\centering
\vspace{-2mm}
\subfigure[\label{psnr_for_loss}]{\includegraphics[scale=0.29]{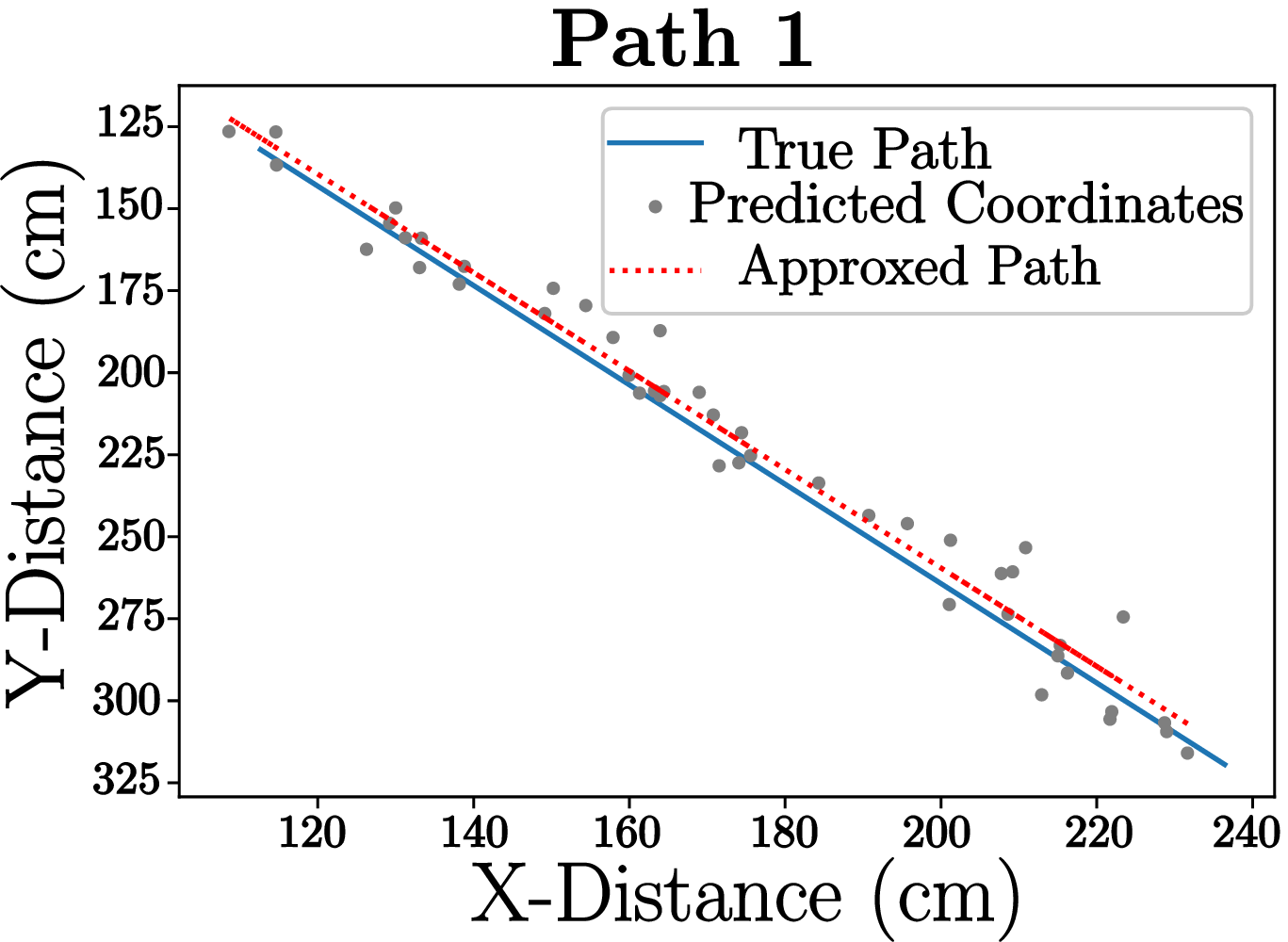}}\hfill
\subfigure[\label{psnr_for_loss_comb}]{\includegraphics[scale=0.29]{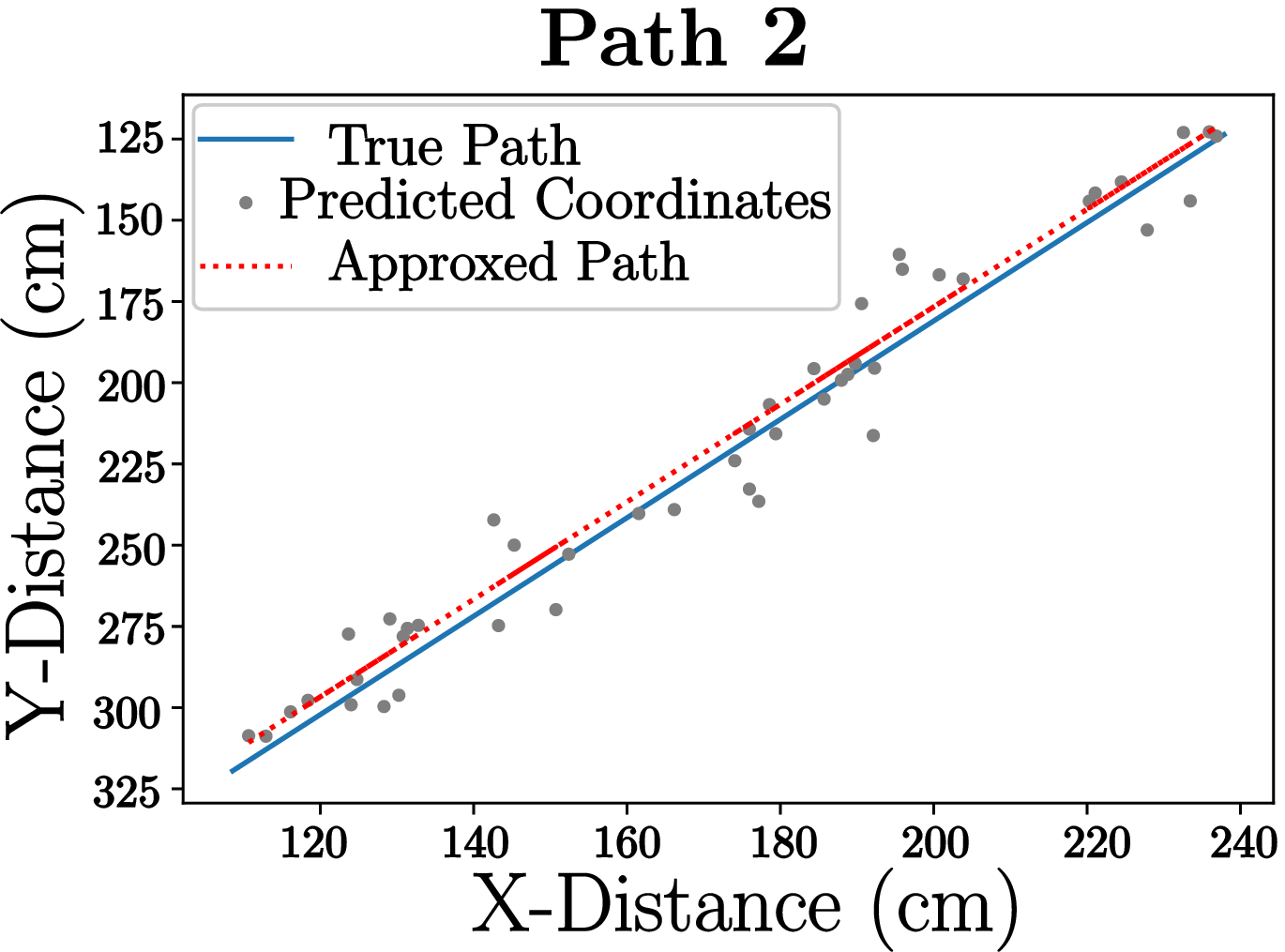}}
\subfigure[\label{psnr_for_loss}]{\includegraphics[scale=0.285]{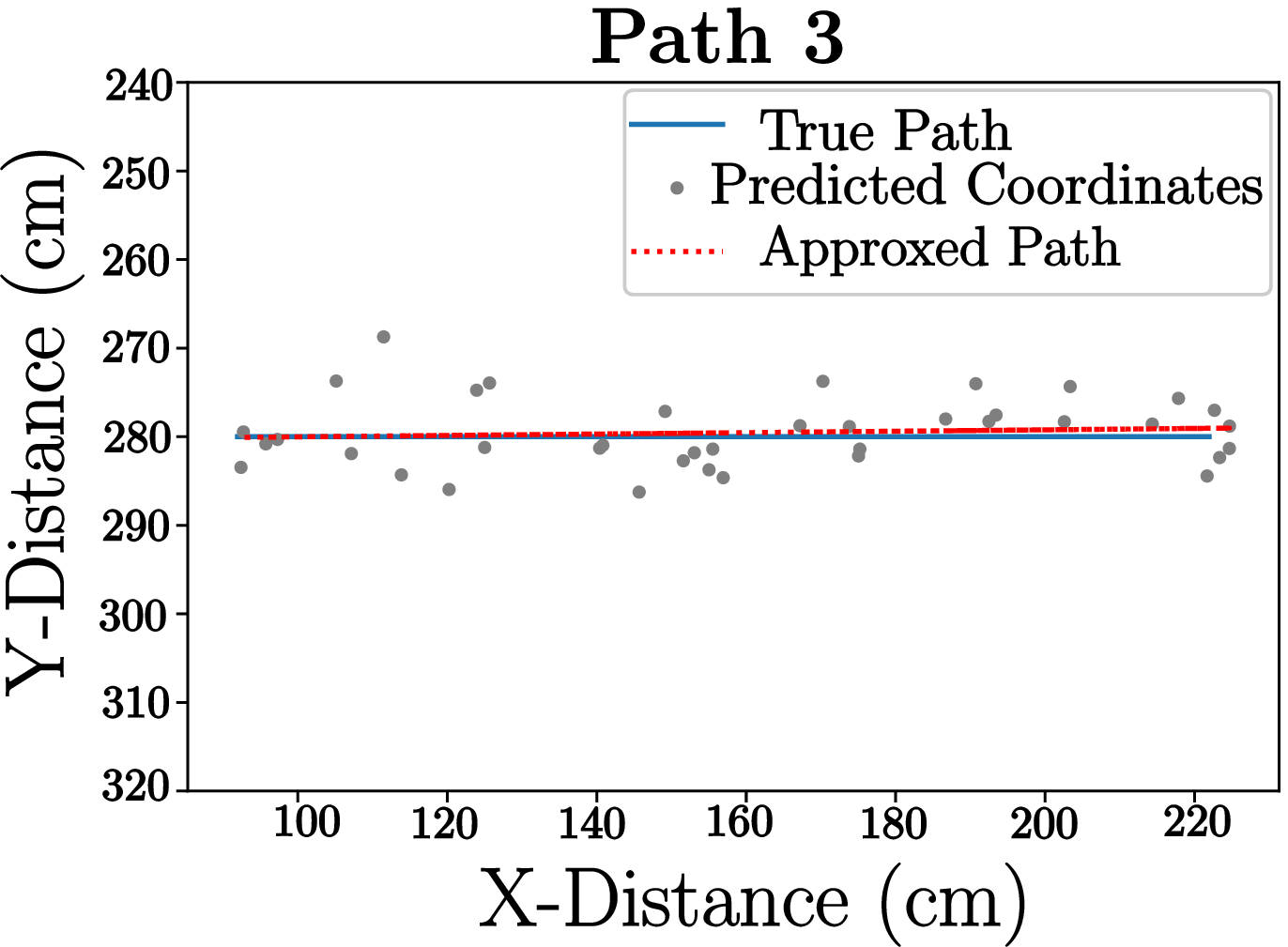}}\hfill
\subfigure[\label{psnr_for_loss_comb}]{\includegraphics[scale=0.29]{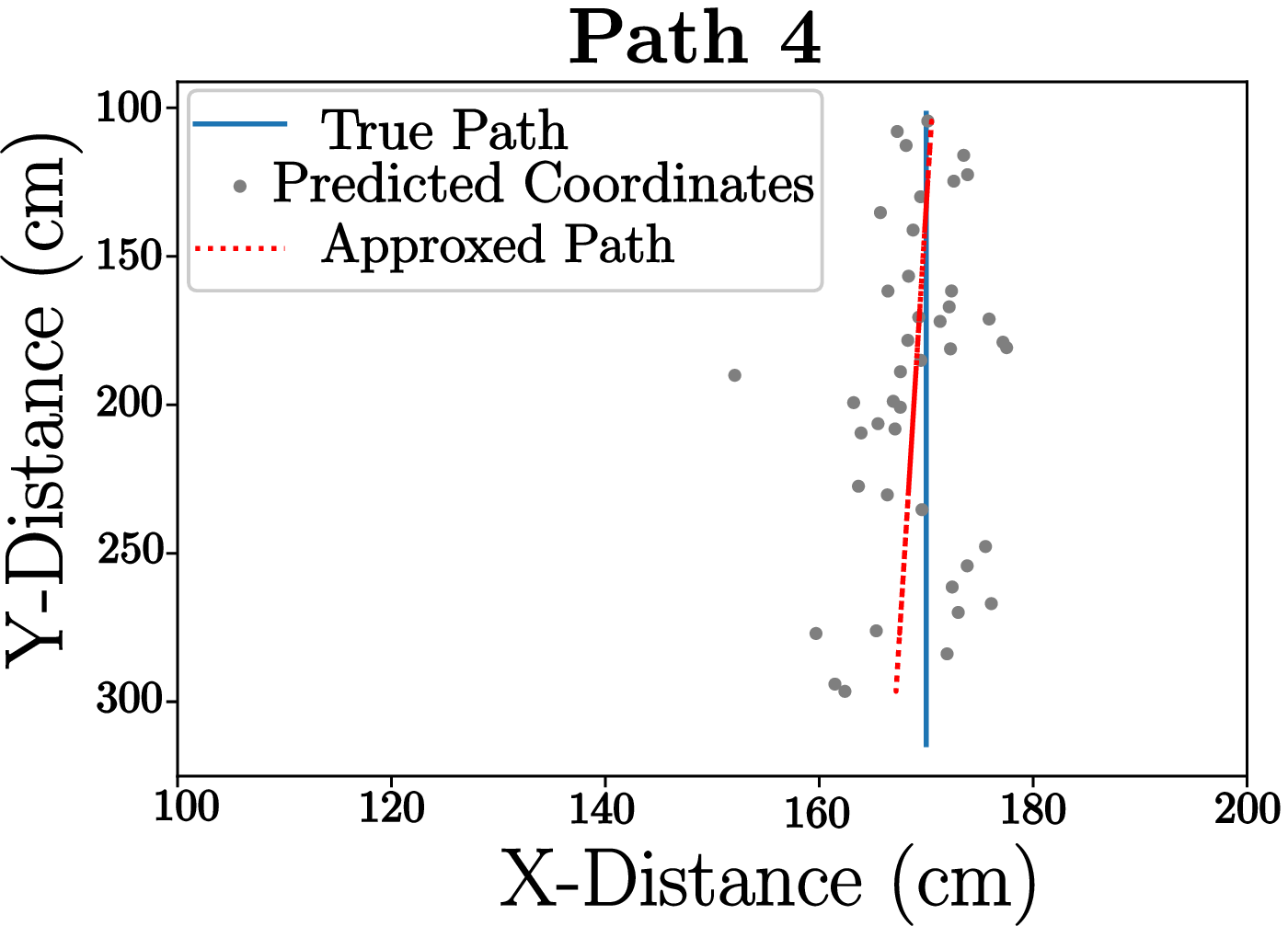}}
\caption{Trajectories predicted by the tracking network, for the four different walking paths defined in Fig. 6. It is evident from the observation that the approximated trajectory (red) is close to the true trajectory (blue) in each instance.}
\label{fig:path}
\vspace{-4mm}
\end{figure}
Fig. \ref{fig:path} illustrates the estimated trajectories for the four different paths of the \textit{walk} activity. The ground truth is represented by the blue lines, whereas the Cartesian coordinates predicted by the network are represented by the small gray circles. The red dotted line illustrates the predicted trajectory, which is the regression line estimated from the predicted Cartesian coordinates using the least squares regression technique. It is evident that the proposed network is able to successfully track the trajectory for the first three paths of the user, including the starting and the end coordinates of the trajectory. We can observe the predicted path diverging from the ground truth in the fourth path, as the user approaches the side of the sensing area where a receiver has not been placed. It is not hard to see that placing a third receiver at $(340,170)$ (please refer Fig. \ref{fig:setup}) will lead to superior performance in all predictions (we were unable to experiment with a three receiver set up due to hardware limitations). 

When comparing the inference time between B-GRU and other uni-directional models, B-GRU inference time is 1.75 times higher than that of the other two models. However, B-GRU completes the inferring within 0.7\% of the sample duration, which qualifies for a real time system.
\setlength\tabcolsep{1.5pt}
\begin{table}[t]
\caption{Error rates for Activity Recognition.}
\centering
\begin{tabular}{ |p{0.76cm}|p{0.76cm}|p{0.76cm}|p{0.76cm}|p{0.76cm}|p{0.76cm}|p{0.76cm}|p{0.76cm}|p{0.76cm}|}
\hline
\textit{sit}&\textit{jump}&\textit{fall}&\textit{run}&\textit{NoAc}&\textit{walk1}&\textit{walk2}&\textit{walk3}&\textit{walk4}\\
\hline
1.65\%&0\%&0.68\%&1.58\%&0\%&1.38\%&1.01\%&2.49\%&1.71\%\\
\hline
\end{tabular}
\label{table:error rate}
\end{table}

\subsection{Performance of the Combined Multi-task Network}
\begin{table}[t]
\vspace{-2mm}
\caption{Results of the combined multi-task network.}
\centering
\begin{tabular}{ |{c}|{c}|{c}|}
\hline
Task & Model performance & Inference time\\
\hline
Activity recognition & 97.36 \% & 14.35 ms \\
\hline
Authentication & 95.90 \% &  14.35 ms\\
\hline
Tracking & $\pm$ 12 cm & 14.35 ms\\
\hline
\end{tabular}
\label{table:combined}
\vspace{-4mm}
\end{table}

We empirically find the values for $\alpha,\beta$ and $\gamma$ which provides the highest performance to be $0.15,0.15$ and $0.70$, respectively. Using these values, we obtain the results tabulated in Table \ref{table:combined}. The performance of the combined model for the three tasks marginally falls below the performance of the respective individual models, yet, achieves a significant speedup. The ensemble models collectively require $5.82+8.64+8.72 = 23.18$ ms for the entire prediction, whereas the combined model requires only $14.35$ ms, achieving a speedup of $38 \%$. Operating all the ensemble models in parallel is not feasible in practice, since the memory constraints of GPUs prevent a large number of models required for ensembling being concurrently loaded, compelling the system to operate each ensemble sequentially.

\subsection{Results of the robustness analysis}
\begin{table}[t]
\caption{Confidence scores of the in-the-wild dataset.}
\centering
\begin{tabular}{ |{c}|p{2cm}|p{1.1cm}|p{1.4cm}|}
\hline
Task & Confidence score margin  & Main test set & In-the-wild dataset\\
\hline
Activity recognition & 75\% & 98.85\% & \textbf{86.86\%} \\
\hline
Authentication & 99.9\% & 79.12\% & \textbf{25.57\%}\\
\hline
\end{tabular}
\label{table:wild}
\end{table}

We base our analysis of robustness of the activity recognizer and the authenticator on the highest confidence score of its prediction, as summarized in Table \ref{table:wild}. In case of the authenticator, we argue that the confidence of a successful authentication should be near perfect and we raise the confidence score margin to a high value of $99.9\%$. This is due to the fact that if an authenticator network recognizes an individual, it should almost be fully confident about its prediction, due to the sensitive nature of authentication. Thus, after thresholding, the accuracy for the in-the-wild dataset should be as low as possible, to avoid unauthorized persons being authenticated. Raising the confidence margin in this manner reduces the performance on the main test set to a $79.12\%$, yet, we can tolerate false negatives over false positives from an authentication system. The authenticator only predicts $25.57\%$ of the in-the-wild dataset, establishing that it often fails to recognize a person that it has never seen before, demonstrating its robustness.

For the activity recognizer, we accept the prediction of the network if the confidence of prediction is above $75\%$. The confidence of a successful prediction should be high, yet intuitively, the threshold need not be enforced as strictly as the authenticator.  Hence, we choose a generic value of 75\% as the confidence threshold. $98.85\%$ of the activity predictions on the main test set by the activity recognizer were made with a confidence score in excess of $75\%$, whereas a significant portion of $86.86\%$ of the activity predictions on the in-the-wild dataset were also made with a confidence score in excess of $75\%$. Hence, it can be concluded that the activity recognizers can successfully recognize the activities of a person that the network has never seen before, which ensures its robustness.

\vspace{-2mm}
\subsection{Potential enhancements to the proposed system}

Even though the simple setup proposed in Section \ref{sec:experiment_setup} provided adequate performance, a superior performance can be obtained via several enhancements. Our system discarded the phase information of the estimated CSI for implementational simplicity, yet integrating phase information for a system where the required additional processing can be dispensed can lead to better performance. Similarly, increasing the number of receivers and the number of antennas per receiver can provide finer predictions in more complex scenarios. Optimal antenna placement for the transmitters and receivers can also enhance the performance of the system.
\vspace{-3mm}
\section{Conclusions} \label{sec:conclusion}
This paper proposed a novel system capable of completely characterize an event, using processed channel state information gathered from commercial Wi-Fi devices. The system performs activity recognition, authentication and tracking simultaneously, by utilizing deep neural networks. It is fully autonomous, non-invasive, requires zero user intervention, device free and passive.
Experimental results were presented to demonstrate the feasibility and the achievable performance of the proposed system. The results have shown that the proposed system achieves promising prediction scores for all three tasks on a collected dataset, only using two single antenna Wi-Fi receivers. The robustness of the system has been established through an in-the-wild study. Possible improvements of the proposed system have also been highlighted. Future work include performance improvements for the single user case and extending the framework to accommodate multiple simultaneous users.


\vspace{-2mm}
%

\begin{thebibliography}{10}
\providecommand{\url}[1]{#1}
\csname url@samestyle\endcsname
\providecommand{\newblock}{\relax}
\providecommand{\bibinfo}[2]{#2}
\providecommand{\BIBentrySTDinterwordspacing}{\spaceskip=0pt\relax}
\providecommand{\BIBentryALTinterwordstretchfactor}{4}
\providecommand{\BIBentryALTinterwordspacing}{\spaceskip=\fontdimen2\font plus
\BIBentryALTinterwordstretchfactor\fontdimen3\font minus
  \fontdimen4\font\relax}
\providecommand{\BIBforeignlanguage}[2]{{%
\expandafter\ifx\csname l@#1\endcsname\relax
\typeout{** WARNING: IEEEtran.bst: No hyphenation pattern has been}%
\typeout{** loaded for the language `#1'. Using the pattern for}%
\typeout{** the default language instead.}%
\else
\language=\csname l@#1\endcsname
\fi
#2}}
\providecommand{\BIBdecl}{\relax}
\BIBdecl

\bibitem{metzler2012appearance}
J.~Metzler, ``Appearance-based re-identification of humans in low-resolution
  videos using means of covariance descriptors,'' in \emph{Proc. IEEE
  International Conference on Advanced Video and Signal-Based Surveillance\em},
  Sep. 2012.

\bibitem{han2019can}
H.~Han, M.~Zhou, and Y.~Zhang, ``Can virtual samples solve small sample size
  problem of kissme in pedestrian re-identification of smart transportation?''
  \emph{IEEE Transactions on Intelligent Transportation Systems}, 2019.

\bibitem{simonyan2014two}
K.~Simonyan and A.~Zisserman, ``Two-stream convolutional networks for action
  recognition in videos,'' in \emph{Advances in neural information processing
  systems}, pp. 568--576, 2014.

\bibitem{kang2003continuous}
J.~Kang, I.~Cohen, and G.~Medioni, ``Continuous tracking within and across
  camera streams,'' in \emph{Proc. IEEE/CVF Conference on Computer Vision and
  Pattern Recognition\em}, Jun. 2003.

\bibitem{yuan2017automatic}
X.~Yuan, L.~Kong, D.~Feng, and Z.~Wei, ``Automatic feature point detection and
  tracking of human actions in time-of-flight videos,'' \emph{IEEE/CAA Journal
  of Automatica Sinica}, vol.~4, no.~4, pp. 677--685, 2017.

\bibitem{isaac2019template}
E.~R. Isaac, S.~Elias, S.~Rajagopalan, and K.~Easwarakumar, ``Template-based
  gait authentication through bayesian thresholding,'' \emph{IEEE/CAA Journal
  of Automatica Sinica}, vol.~6, no.~1, pp. 209--219, 2019.

\bibitem{more2018gait}
S.~A. More and P.~J. Deore, ``Gait recognition by cross wavelet transform and
  graph model,'' \emph{IEEE/CAA Journal of Automatica Sinica}, vol.~5, no.~3,
  pp. 718--726, 2018.

\bibitem{sun2018accelerometer}
F.~Sun, C.~Mao, X.~Fan, and Y.~Li, ``Accelerometer-based speed-adaptive gait
  authentication method for wearable iot devices,'' \emph{IEEE Internet of
  Things Journal}, vol.~6, no.~1, pp. 820--830, 2018.

\bibitem{zeng2016wiwho}
Y.~Zeng, P.~H. Pathak, and P.~Mohapatra, ``Wiwho: Wifi-based person
  identification in smart spaces,'' in \emph{Proc. IEEE International
  Conference on Information Processing in Sensor Networks}, p.~4, 2016.

\bibitem{zhang2016wifi}
J.~Zhang, B.~Wei, W.~Hu, and S.~S. Kanhere, ``Wifi-id: Human identification
  using wifi signal,'' in \emph{Proc. IEEE International Conference on
  Distributed Computing in Sensor Systems}, pp. 75--82, 2016.

\bibitem{pokkunuru2018neuralwave}
A.~Pokkunuru, K.~Jakkala, A.~Bhuyan, P.~Wang, and Z.~Sun, ``Neuralwave:
  Gait-based user identification through commodity wifi and deep learning,'' in
  \emph{Annual Conference of the IEEE Industrial Electronics Society}, pp.
  758--765, 2018.

\bibitem{lin2018wiau}
C.~Lin, J.~Hu, Y.~Sun, F.~Ma, L.~Wang, and G.~Wu, ``Wiau: An accurate
  device-free authentication system with resnet,'' in \emph{Proc. IEEE
  International Conference on Sensing, Communication, and Networking}, pp.
  1--9, 2018.

\bibitem{pu2013whole}
Q.~Pu, S.~Gupta, S.~Gollakota, and S.~Patel, ``Whole-home gesture recognition
  using wireless signals,'' in \emph{Proc. ACM International Conference on
  Mobile Computing \& networking\em}, Sep. 2013.

\bibitem{wang2016we}
G.~Wang, Y.~Zou, Z.~Zhou, K.~Wu, and L.~M. Ni, ``We can hear you with
  {W}i-{F}i!'' \emph{IEEE Transactions on Mobile Computing}, vol.~15, no.~11,
  pp. 2907--2920, Nov. 2016.

\bibitem{yun2017strata}
S.~Yun, Y.-C. Chen, H.~Zheng, L.~Qiu, and W.~Mao, ``Strata: Fine-grained
  acoustic-based device-free tracking,'' in \emph{Proc. ACM Annual
  International Conference on Mobile Systems, Applications, and Services}, pp.
  15--28, 2017.

\bibitem{yousefi2017survey}
S.~Yousefi, H.~Narui, S.~Dayal, S.~Ermon, and S.~Valaee, ``A survey on behavior
  recognition using {W}i-{F}i channel state information,'' \emph{IEEE
  Communications Magazine}, vol.~55, no.~10, pp. 98--104, Oct. 2017.

\bibitem{wang2016csi}
X.~Wang, L.~Gao, S.~Mao, and S.~Pandey, ``Csi-based fingerprinting for indoor
  localization: A deep learning approach,'' \emph{IEEE Transactions on
  Vehicular Technology}, vol.~66, no.~1, pp. 763--776, 2016.

\bibitem{kim2018scalable}
K.~S. Kim, S.~Lee, and K.~Huang, ``A scalable deep neural network architecture
  for multi-building and multi-floor indoor localization based on wi-fi
  fingerprinting,'' \emph{Big Data Analytics}, vol.~3, no.~1, p.~4, 2018.

\bibitem{jang2018indoor}
J.-W. Jang and S.-N. Hong, ``Indoor localization with wifi fingerprinting using
  convolutional neural network,'' in \emph{Proc. IEEE International Conference
  on Ubiquitous and Future Networks}, pp. 753--758, 2018.

\bibitem{youssef2007challenges}
M.~Youssef, M.~Mah, and A.~Agrawala, ``Challenges: device-free passive
  localization for wireless environments,'' in \emph{Proc. ACM International
  Conference on Mobile Computing and Networking\em}, Oct. 2007.

\bibitem{li2017indotrack}
X.~Li, D.~Zhang, Q.~Lv, J.~Xiong, S.~Li, Y.~Zhang, and H.~Mei, ``Indotrack:
  Device-free indoor human tracking with commodity {W}i-{F}i,'' vol.~1, no.~3,
  p.~72, Sep. 2017.

\bibitem{adib20143d}
F.~Adib, Z.~Kabelac, D.~Katabi, and R.~C. Miller, ``3{D} tracking via body
  radio reflections,'' in \emph{Proc. USENIX Symposium on Networked Systems
  Design and Implementation\em}, Apr. 2014.

\bibitem{wang2016device}
J.~Wang, X.~Zhang, Q.~Gao, H.~Yue, and H.~Wang, ``Device-free wireless
  localization and activity recognition: A deep learning approach,'' \emph{IEEE
  Transactions on Vehicular Technology}, vol.~66, no.~7, pp. 6258--6267, 2016.

\bibitem{li2016robust}
F.~Li, M.~Al-qaness, Y.~Zhang, B.~Zhao, and X.~Luan, ``A robust and device-free
  system for the recognition and classification of elderly activities,''
  \emph{IEEE Sensors}, vol.~16, no.~12, p. 2043, Dec. 2016.

\bibitem{zhang2019wifimap+}
W.~Zhang, S.~Zhou, L.~Yang, L.~Ou, and Z.~Xiao, ``Wifimap+: High-level indoor
  semantic inference with wifi human activity and environment,'' \emph{IEEE
  Transactions on Vehicular Technology}, 2019.

\bibitem{li2006orthogonal}
Y.~G. Li and G.~L. Stuber, \emph{Orthogonal frequency division multiplexing for
  wireless communications}, 1st~ed.\hskip 1em plus 0.5em minus 0.4em\relax
  \hspace{-2.1mm} Springer Science \& Business Media, 2006.

\bibitem{yang2013rssi}
Z.~Yang, Z.~Zhou, and Y.~Liu, ``From {RSSI} to {CSI}: Indoor localization via
  channel response,'' \emph{ACM Computing Surveys}, vol.~46, no.~2, p.~25, Nov.
  2013.

\bibitem{sen2012you}
S.~Sen, B.~Radunovic, R.~R. Choudhury, and T.~Minka, ``You are facing the
  {M}ona {L}isa: {S}pot localization using {PHY} layer information,'' in
  \emph{Proc. ACM International Conference on Mobile Systems, Applications, and
  Services\em}, Jun. 2012.

\bibitem{wang2015understanding}
W.~Wang, A.~X. Liu, M.~Shahzad, K.~Ling, and S.~Lu, ``Understanding and
  modeling of {W}i-{F}i signal based human activity recognition,'' in
  \emph{Proc. ACM International Conference on Mobile Computing and
  Networking\em}, Sep. 2015.

\bibitem{srivastava2014dropout}
N.~Srivastava, G.~Hinton, A.~Krizhevsky, I.~Sutskever, and R.~Salakhutdinov,
  ``Dropout: a simple way to prevent neural networks from overfitting,''
  \emph{The journal of machine learning research}, vol.~15, no.~1, pp.
  1929--1958, 2014.

\bibitem{huang2017densely}
G.~Huang, Z.~Liu, L.~Van Der~Maaten, and K.~Q. Weinberger, ``Densely connected
  convolutional networks,'' in \emph{Proceedings of the IEEE conference on
  computer vision and pattern recognition}, pp. 4700--4708, 2017.

\bibitem{graves2013speech}
A.~Graves, A.-r. Mohamed, and G.~Hinton, ``Speech recognition with deep
  recurrent neural networks,'' in \emph{2013 IEEE international conference on
  acoustics, speech and signal processing}, pp. 6645--6649.\hskip 1em plus
  0.5em minus 0.4em\relax IEEE, 2013.

\bibitem{nwankpa2018activation}
C.~Nwankpa, W.~Ijomah, A.~Gachagan, and S.~Marshall, ``Activation functions:
  Comparison of trends in practice and research for deep learning,''
  \emph{arXiv preprint arXiv:1811.03378}, 2018.

\bibitem{kingma2014adam}
D.~P. Kingma and J.~Ba, ``Adam: A method for stochastic optimization,''
  \emph{arXiv preprint arXiv:1412.6980}, 2014.

\bibitem{dietterich2000ensemble}
T.~G. Dietterich, ``Ensemble methods in machine learning,'' in \emph{Proc.
  Springer International workshop on multiple classifier systems\em}, Jun.
  2000.

\end{thebibliography}

\end{document}